\documentstyle[11pt,epsf]{article}
\newcommand{\be}{\begin{equation}}
\newcommand{\ee}{\end{equation}} 
\newcommand{\eei}{\end{equation}\indent\indent}
\newcommand{\bc}{\begin{center}}
\newcommand{\ec}{\end{center}}
\newcommand{\ber}{\begin{eqnarray}}
\newcommand{\ear}{\end{eqnarray}}
\newcommand{\ba}{\begin{array}}
\newcommand{\ea}{\end{array}}

\def\case#1/#2{\textstyle\frac{#1}{#2} }
\begin{document}
\title{Autocatalytic Theory of Meaning}
\author{Mark D. Roberts, \\\\
Department of Mathematics and Applied Mathematics,\\
University of Cape Town,\\
Rondbosch 7701,\\
South Africa\\\\
roberts@gmunu.mth.uct.ac.za} 
\date{\today}
\maketitle
\bc Eprint number http://xxx.lanl.gov/abs/cs.CL/9902027 \ec
\bc Mathematical Reviews Subject Classification: \ec
\bc http://www.ams.org/msc/ \ec
\bc  92J10,  92H10 \ec
\bc Association for Computing Machinery Classification:  \ec
\bc http://www.acm.org/class/1998/overview.html \ec
\bc  I.2.6;  J.4;  I.2.7  \ec 
\newline
\begin{abstract}
Recently it has been argued that autocatalytic
theory could be applied to the origin of culture.
Here possible application to a theory of meaning
in the philosophy of language,  called radical interpretation,
is commented upon and compared to previous applications. 
\end{abstract}
\section{Autocatalytic theory of the Origin of Life.}
Gabora (1998) \cite{bi:gabora} discusses the autocatalytic theory of
the origin of life in sections 26-32.   Instead of having
one self-replicating molecule,  there are a set of molecules
each of which can replicate a different member of the set.
The appearance of such a set may or may not be more likely
than the appearance of a single self-replicating molecule.
Also such a set may or may not evolve at a fast rate 
conditioned by selection.
\section{Autocatalytic theory of the Origin of Culture.}
From sections 33-60 Gabora investigates whether a 
similar mechanism could explain the origin of 
human culture.   There are various aspects of human 
culture the origin of which calls for explanation.
The central problem is to explain how the mind 
transforms from being episodic,  or only being
capable of recalling episodes,  to being capable
of abstraction.   Once the ability to abstract 
is present it will be selected for and increase,  
because abstraction allows for creative acts.   
The transition to an abstracting mind could originate 
by happening once and then developing - this is
the analogy of the self-replicating molecule;  or
there could be several ideas occurring at once 
which by chance help in problem solving - this is
the analog of a set of molecules each of which 
replicate a different member of the set.
\section{Autocatalytic approach to Language Acquisition.}
One may ask how many other topics this mechanism
can be applied to.   There are several other places
where the origin of thing is unclear and development 
is fast.   Unclear origin and fast development are 
two of the facets that autocatalytic theory might 
help to explain.   This suggests application to
language acquisition Pinker (1984) \cite{bi:pinker}.
\section{Radical Interpretation.}
Radical interpretation,  is a part of the philosophy 
of language concerned with giving an account of how
language has meaning.   The topic has been
developed by Davidson and the original papers collected 
together in a book Davidson (1984) \cite{bi:davidson}.   
The main idea is that one can establish truth or otherwise
of sentences by means external to the spoken 
or written language.   Having established 
truth of a sentence one can gather what it 
means again by factors external to the spoken
or written language.   I have compared this
process to biological evolution,  Roberts (1998) \cite{bi:roberts}.
One of the areas of discussion is how large a
structure radical interpretation should apply
to.   Lewis (1974) \cite{bi:lewis} and Roberts (1998) \cite{bi:roberts}
argue that it should be applied to more than just 
language,  perhaps to any social structure.
Taking this to an extreme would be to apply
it to all of culture:  Gabora's work can be 
viewed as embodying this view.
\section{Autocatalytic approach to Radical Interpretation.}
One can ask what would be the autocatalytic
approach to radical interpretation and how
it would work.   Instead of truth and meaning 
being assigned to sentences individually
they would be assigned to several sentences
concurrently.   In some ways this is preferable
to the standard view as it does not tie meaning
down to the specific wording of sentences.
One would hope to be able to illustrate how 
this mechanism would work using the specific 
test sentences used repeatedly in the philosophy 
of language such as "snow is white' and "grass is green",  
but it is not clear how this could be achieved.
Perhaps the best way of looking at whether an autocatalytic
approach is appropriate is to consider what would happen
if meaning was {\bf not} autocatalytic in nature.
In this case it would be possible to assign meaning
to sentences individually.   In other words given
a fixed sentence - say "snow is white" it would be
possible to assign meaning to it without reference
to any other sentence.   This position is absurd 
as it does not allow for the creativity of language,
where entirely new sentences are continuously being
uttered for the first time.   This {\it reducio ad absurdum}
suggests that meaning is indeed assigned is some autocatalytic 
manner.
\section{Acknowledgement}
This work has been supported by the South African 
Foundation for Research and Development (FRD).

\end{document}